\documentclass{article}

\PassOptionsToPackage{numbers, compress}{natbib}

    \usepackage[preprint]{neurips_2025}

\usepackage[utf8]{inputenc} 
\usepackage[T1]{fontenc}    
\usepackage{hyperref}       
\usepackage{url}            
\usepackage{booktabs}       
\usepackage{amsfonts}       
\usepackage{nicefrac}       
\usepackage{microtype}      
\usepackage{xcolor}         
\usepackage{microtype}
\usepackage{graphicx}
\usepackage{colortbl}

\usepackage{amsmath}
\usepackage{amssymb}
\usepackage{mathtools}
\usepackage{amsthm}
\usepackage{multirow,hhline}
\usepackage{caption}

\usepackage{multirow}
\usepackage{pifont}
\usepackage{makecell}
\usepackage{wrapfig}
\usepackage{subcaption}

\usepackage{enumitem}

\usepackage[capitalize,noabbrev]{cleveref}

\usepackage{graphics}
\usepackage[export]{adjustbox}
\usepackage{bbding}

\usepackage{enumitem}

\usepackage[textsize=tiny]{todonotes}

\newcommand{\model}{\textit{SegEarth-R1}}

\newcommand{\eg}{\textit{e.g.}}
\newcommand{\ie}{\textit{i.e.}}
\newcommand{\etc}{\textit{etc.}}

\newlength\savewidth

\title{SegEarth-R1: Geospatial Pixel Reasoning via Large Language Model}

\makeatletter
\def\thanks#1{\protected@xdef\@thanks{\@thanks
        \protect\footnotetext{#1}}}
\makeatother

\author{
  \vspace{-25pt}\\
  \textbf{Kaiyu Li$^{1,*}$,
  \quad Zepeng Xin$^{1,*}$,
  \quad Li Pang$^1$,
  \quad Chao Pang$^{2}$,
  \quad Yupeng Deng$^{3}$} \\
  \textbf{ Jing Yao$^{3}$,
  \quad Guisong Xia$^{2}$,
  \quad Deyu Meng$^1$,
  \quad Zhi Wang$^1$,
  \quad Xiangyong Cao$^{1,\dag}$}\\
  $^1$Xi'an Jiaotong University ~~\quad $^2$Wuhan University ~~\quad $^3$Chinese Academy of Sciences\\
  \vspace{-25pt}
}
  
\definecolor{mygray}{gray}{.91}

\begin{document}

\maketitle


\renewcommand{\thefootnote}{*}
\footnotetext[1]{Equal contribution}
\renewcommand{\thefootnote}{\dag}
\footnotetext[1]{Corresponding author: caoxiangyong@mail.xjtu.edu.cn}
\renewcommand{\thefootnote}{\arabic{footnote}}


\begin{abstract}
Remote sensing has become critical for understanding environmental dynamics, urban planning, and disaster management. However, traditional remote sensing workflows often rely on explicit segmentation or detection methods, which struggle to handle complex, implicit queries that require reasoning over spatial context, domain knowledge, and implicit user intent. Motivated by this, we introduce a new task, \ie, geospatial pixel reasoning, which allows implicit querying and reasoning and generates the mask of the target region. To advance this task, we construct and release the first large-scale benchmark dataset called EarthReason, which comprises 5,434 manually annotated image masks with over 30,000 implicit question-answer pairs. Moreover, we propose SegEarth-R1, a simple yet effective language-guided segmentation baseline that integrates a hierarchical visual encoder, a large language model (LLM) for instruction parsing, and a tailored mask generator for spatial correlation. The design of SegEarth-R1 incorporates domain-specific adaptations, including aggressive visual token compression to handle ultra-high-resolution remote sensing images, a description projection module to fuse language and multi-scale features, and a streamlined mask prediction pipeline that directly queries description embeddings. Extensive experiments demonstrate that SegEarth-R1 achieves state-of-the-art performance on both reasoning and referring segmentation tasks, significantly outperforming traditional and LLM-based segmentation methods. Our data and code will be released at \url{https://github.com/earth-insights/SegEarth-R1}.



\end{abstract}

\begin{figure}[t]
  \centering
   \includegraphics[width=1.0\linewidth]{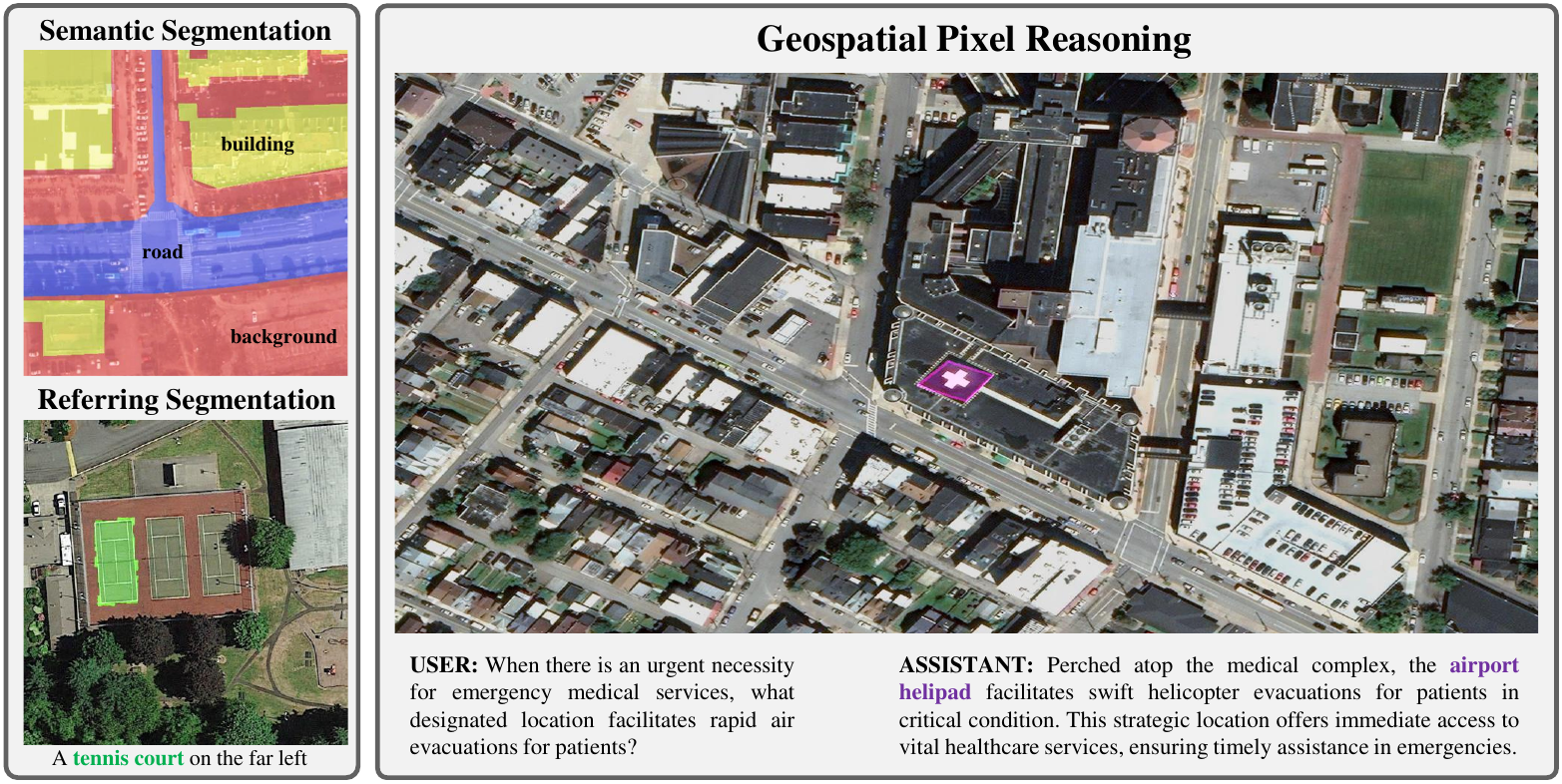}
   \caption{\small Comparison of semantic segmentation, referring segmentation and geospatial pixel inference. (left) Samples from the LoveDA~\cite{wang2021loveda} and RRSIS-D~\cite{liu2024rotated} datasets. (right) Samples from the EarthReason dataset. Previous tasks are limited by fixed taxonomies and explicit instructions, while geospatial pixel reasoning supports complex implicit instructions and requires the reasoning capability of the model.}
   \label{fig:data_sample}
   \vspace{-1em}
\end{figure}

\section{Introduction}
\label{sec:intro}

Earth observation through remote sensing has emerged as a cornerstone of modern geospatial analysis, enabling unprecedented insights into environmental dynamics, urban planning, and disaster management~\cite{rolf2024mission, lu2025vision}. Satellite and aerial images provide a unique vantage point for monitoring planetary-scale phenomena, ranging from deforestation patterns to coastal erosion. However, converting this raw pixel data into actionable insights requires more than traditional computer vision techniques; it demands models capable of reasoning about spatial context, domain knowledge, and implicit user intent. Conventional remote sensing workflows predominantly rely on explicit tasks, \eg, semantic segmentation and referring segmentation~\cite{long2021creating, christie2018functional, yuan2024rrsis}, which operate within fixed taxonomies and require precise user instructions. While effective for well-defined scenarios, these approaches struggle to accommodate complex, implicit queries—for example, identifying regions at elevated risk of landslides based on slope, vegetation cover, and proximity to infrastructure. Such tasks limit implicit reasoning over heterogeneous spatial patterns, object relationships, and environmental metadata, exceeding the capabilities of standard segmentation or detection pipelines.

Motivated by this, we introduce a new task, \ie, geospatial pixel reasoning, which allows implicit querying and reasoning and generates the mask of the target region. To enable research in this task, we build and release the first large-scale benchmark dataset, called EarthReason, which contains 5,434 manually annotated remote sensing image-mask pairs drawn from diverse classification sources, covering 28 scene categories at spatial resolutions ranging from 0.5m to 153m. Each image is paired with multiple implicit reasoning questions that require the model to infer target masks based on contextual and domain-specific knowledge, rather than explicit object names. In addition, by incorporating empty target cases and varying spatial scales, EarthReason pushes models to generalize across complex real-world remote sensing scenarios.


Recent progress in multimodal large language models (MLLMs) has demonstrated impressive performance in natural image domains, where models like LISA~\cite{lai2024lisa} and PixelLM~\cite{ren2024pixellm} leverage large language models (LLMs)~\cite{touvron2023llama, chiang2023vicuna, yang2024qwen2} to interpret rich textual prompts and generate pixel-level outputs. These frameworks excel at tasks such as reasoning segmentation~\cite{lai2024lisa}, where the target mask is not directly specified but must be inferred from nuanced language cues. Unfortunately, directly transferring these methods to geospatial pixel reasoning is non-trivial since remote sensing images present extreme scale variation, densely packed small-scale objects and ultra-high resolution that violate assumptions of natural images. Moreover, different from natural images, remote sensing queries often require spatial correlations. For instance, identifying ``informal settlements'' relies on detecting roof material irregularities, road network fragmentation, and spatial adjacency to legal land-use zones.

To address these challenges, we present SegEarth-R1, a simple yet effective language-guided segmentation model that integrates a hierarchical visual encoder, an LLM for instruction parsing, and a tailored mask generator designed for spatial correlation. Further, some components are also designed to adapt to the characteristics of remote sensing images. Specifically, we propose the aggressive visual token compression to handle ultra-high-resolution images, a description projection module to fuse language and multi-scale features, and a streamlined mask prediction pipeline that directly queries description embeddings. Despite its architectural simplicity, SegEarth-R1 achieves advanced performance on EarthReason and referring segmentation datasets, significantly outperforming both traditional and LLM-based segmentation methods.

In summary, our contributions are as follows:
\begin{itemize}[leftmargin=8pt]
\item We introduce the geospatial pixel reasoning task, which requires models to infer segmentation masks from implicit natural language queries by reasoning over spatial context and domain knowledge.
\item We build and release the first large-scale benchmark with 5,434 image-mask pairs, 28 categories, and over 30,000 implicit question-answer pairs, fostering research in geospatial pixel reasoning.
\item We propose an LLM-based segmentation model, SegEarth-R1, which incorporates new segmentation capabilities in remote sensing, containing several domain-specific designs.
\item Extensive experiments show that SegEarth-R1 achieves state-of-the-art performance on reasoning and referring segmentation tasks, compared to traditional methods and other LLM-based methods.

\end{itemize}

\section{Related Work}
\label{sec:related_work}



\subsection{Referring Segmentation}

Referring segmentation aims to segment targets in an image based on natural language descriptions, requiring precise alignment between linguistic expressions and visual content. Early approaches adopted CNN-RNN/LSTM frameworks~\cite{hu2016segmentation, liu2017recurrent, li2018referring, margffoy2018dynamic, shi2018key, huang2020referring} to extract visual features and encode textual queries, respectively. However, these methods struggled with complex expressions due to limited local receptive fields and insufficient cross-modal interaction~\cite{ji2024survey}. To address these limitations, attention mechanisms~\cite{vaswani2017attention} emerged as a pivotal technique~\cite{ding2021vision, yang2022lavt, wu2022language, hu2023beyond, xu2023bridging, nag2024safari, wu2024towards, shang2024prompt}. VLT~\cite{ding2021vision} dynamically generates adaptive query vectors based on image-text interactions, enabling precise localization through cross-modal attention. LAVT~\cite{yang2022lavt} further advances this paradigm by integrating hierarchical visual-linguistic fusion within a Swin Transformer~\cite{liu2021swin} backbone, where pixel-word attention refines multiscale features to achieve fine-grained semantic alignment. In remote sensing, specifying segmentation for certain instances can improve interpretation efficiency and user interactivity. Recently, Yuan \textit{et al.}~\cite{yuan2024rrsis} introduced referring segmentation into satellite images for the first time. Subsequently, following the LAVT~\cite{yang2022lavt} architecture, RMSIN~\cite{liu2024rotated} also incorporated adaptive rotated convolutions to address scale and orientation variations. FIANet~\cite{lei2024exploring} and CroBIM~\cite{dong2024cross} introduced elaborate cross-modal interactions for feature alignment. RSSep~\cite{ho2024rssep} reformulated referring segmentation as a sequence-to-sequence task, predicting polygonal boundaries to handle scale variations and blurred edges~\cite{liu2023polyformer}. However, existing methods effectively follow explicit instructions for target segmentation but lack implicit intent reasoning. In this paper, the proposed geospatial pixel reasoning task advances beyond referring segmentation by employing LLMs' reasoning capabilities to interpret subtle instructions and accurately segment desired targets.

\subsection{LLM-based Segmentation}

Recent advances in LLMs have significantly expanded their capabilities to integrate pixel-level segmentation with language reasoning~\cite{xiao2023florence, wang2023visionllm, wu2025visionllm, beyer2024paligemma, steiner2024paligemma, zhang2024next, yuan2025sa2va, he2024multi}. For instance, Florence-2~\cite{xiao2023florence} unified text, detection, and segmentation through a sequence-to-sequence framework with task instructions. To address the complexity of real-world segmentation scenarios, some works focus on architectural specialization and instruction-aware adaptation. LISA~\cite{lai2024lisa, yang2023lisa++} established the paradigm by introducing a \texttt{[SEG]} token to connect LLMs with segmentation decoders like SAM~\cite{kirillov2023segment}, enabling language-guided mask prediction. Subsequent studies enhanced this paradigm: GSVA~\cite{xia2024gsva} introduced shared-weight \texttt{[SEG]} tokens and \texttt{[REJ]} tokens for multi-target and empty-target handling~\cite{liu2023gres, ren2024pixellm, zhang2024groundhog}, while GLaMM~\cite{rasheed2024glamm} achieved pixel-grounded conversational capabilities through holistic segmentation~\cite{zhou2024instruction}. Parallel efforts focused on architectural unification - PSALM~\cite{zhang2024psalm} established a flexible input schema for multi-task segmentation, and OMG-LLaVA~\cite{zhang2025omg} combined universal segmentation backbones with LLMs for pixel-level reasoning. Video understanding extensions emerged through VISA~\cite{yan2024visa} and InstructSeg~\cite{wei2024instructseg}, which integrated temporal reasoning. Notably, Text4Seg~\cite{lan2024text4seg} redefined segmentation as a text generation problem using semantic descriptors, eliminating the need for an additional decoder. In remote sensing, benefiting from the above paradigms~\cite{lai2024lisa, lan2024text4seg}, some unified models such as RSUniVLM~\cite{liu2024rsunivlm}, GeoGround~\cite{zhou2024geoground} and GeoPix~\cite{ou2025geopix} are equipped with segmentation capabilities. Although based on LLM, these models focus only on explicit text-guided segmentation. Further, GeoPixel~\cite{shabbir2025geopixel} introduced grounded conversation generation~\cite{rasheed2024glamm} to remote sensing, but it still does not provide reasoning capability. Our SegEarth-R1 also follows the LLM-based segmentation paradigm, but is different from previous methods. Specifically, SegEarth-R1 is the first work to support reasoning about the target region from implicit queries, and its components are specifically designed for the challenges in remote sensing.


\section{Benchmark Geospatial Pixel Reasoning Dataset---EarthReason}

\begin{table*}
  \caption{Comparison between EarthReason and other related datasets. The {\color{white!50!black}gray} rendering denotes the natural image dataset. ``Seg'', ``Det'', ``VG'', ``Cls'' denote segmentation, detection, visual grounding and classification datasets, respectively.}
  \label{table_data}
  \centering
  \scalebox{0.63}{
  \begin{tabular}{@{}lcccccccccc@{}}
    \toprule[1pt]
    Dataset & Mask Label & Reasoning Query & Spatial resolution & Image Size & Image Num & Image Source & Class Num \\
    \midrule[1pt]
    {\color{white!50!black}ReasonSeg} \cite{lai2024lisa} & {\color{white!50!black}\checkmark} & {\color{white!50!black}\checkmark} & {\color{white!50!black}-} & {\color{white!50!black}-} & {\color{white!50!black}1,218} & \makecell[c]{{\color{white!50!black}OpenImages (Seg) \&} \\ {\color{white!50!black}ScanNetv2 (Seg)}} & {\color{white!50!black}-} \\
    {\color{white!50!black}LLM-Seg40K} \cite{wang2024llm} & {\color{white!50!black}\checkmark} & {\color{white!50!black}\checkmark} & {\color{white!50!black}-} & {\color{white!50!black}-} & {\color{white!50!black}14,000} & \makecell[c]{{\color{white!50!black}LVIS (Seg) \&} \\ {\color{white!50!black}EgoObjects (Seg)}} & {\color{white!50!black}-} \\
    \midrule[1pt]
    EarthVQA~\cite{wang2024earthvqa} & \ding{55} & \checkmark & 0.3m & $1024^2$ & 6,000 & LoveDA (Seg) & 14 \\
    RegSegRS~\cite{yuan2024rrsis} & \checkmark & \ding{55} & 0.5m-30m & $800^2$ & 4,420 & SkyScapes (Seg) & 14 \\
    RRSIS-D~\cite{liu2024rotated} & \checkmark & \ding{55} & 0.13m & $512^2$ & 17,402 & RSVGD (VG) \& DIOR (OD) & 20 \\
    RISBench~\cite{dong2024cross} & \checkmark & \ding{55} & 0.1m-30m & $512^2$ & 52,472 & DOTAv2(OD) \& DIOR (OD) & 26 \\
    \midrule
    EarthReason & \checkmark & \checkmark & 0.5m-153m & $123^2$-$7617^2$ & 5,434 & AID (Cls) \& fMoW (Cls) & 28  \\
    \bottomrule[1pt]
  \end{tabular}}
\end{table*}


\subsection{Comparison with Related Dataset}

We analyze three types of tasks and datasets related to geospatial pixel reasoning, \ie, natural image reasoning segmentation, remote sensing visual question answering (VQA), and remote sensing referring segmentation, as shown in Table~\ref{table_data}.
RefSegRS~\cite{yuan2024rrsis} and RRSIS-D~\cite{liu2024rotated} provide early benchmarks with image-text-mask triplets. RISBench~\cite{dong2024cross}, the largest RRSIS dataset to date, introduced 52,472 triplets with oriented bounding boxes and pixel-level masks generated via a semi-automatic pipeline. These datasets address the limitations of earlier text-focused datasets (\eg, RSICD~\cite{lu2017exploring}, EarthVQA~\cite{wang2024earthvqa}, \etc) and enable comprehensive evaluation of multimodal models. Compared to the previous referring segmentation datasets, our EarthReason datasets has the following features: \textbf{(1)} The mask labels in EarthReason are not explicitly specified by the query, but require further reasoning to determine the target, which challenges the model's reasoning ability. \textbf{(2)} EarthReason uses a more raw data source. The previous related datasets directly transform existing segmentation datasets~\cite{azimi2019skyscapes, wang2021loveda} or SAM-processed detection datasets~\cite{zhan2023rsvg, li2020object, ding2021object}, while our EarthReason uses images from classification datasets~\cite{long2021creating, christie2018functional} and we manually annotate them. This allows EarthReason to provide more data gain when it comes to co-training of unified segmentation tasks. \textbf{(3)} EarthReason has more diverse spatial resolutions and image sizes, which are conducive to solving the object scale spanning problem inherent in remote sensing images~\cite{rolf2024mission}. Compared to the first natural image reasoning segmentation dataset, ReasonSeg, EarthReason contains $4.46\times$ more data than it. Therefore, we believe that EarthReason, as the first geospatial pixel reasoning dataset in the remote sensing area, is capable of performing initial explorations of this task.

\subsection{Dataset Generation Pipeline}
Our benchmark dataset EarthReason is generated according to the following three steps, i.e., image collection, question-answer pair generation, and object mask labeling.

\noindent
\textbf{Image Collection.} As mentioned above, to avoid potential data leakage in the future construction of unified segmentation models for remote sensing, we collect images from existing classification data. Although this increases the annotation cost, it also motivates more diverse scenes. Specifically, we first select the 28 categories that are more suitable for reasoning in the Million-AID~\cite{long2021creating} dataset, and sample about 200 images for each category. Then, we find that the actual geographic range contained in Million-AID's images is limited. Thus, we also collect 800 images in the fMoW~\cite{christie2018functional} dataset to enhance the model's reasoning ability in complex scenes. Further, to alleviate the factitious illusion issue~\cite{pang2024vhm}, we add an extra 200 empty target images (\ie, the implied target is not in the image). Finally, some low-quality images are eliminated, and we obtain a total of 5,434 images.

\noindent
\textbf{Question-Answer Pair Generation.} We use GPT-4o\footnote{\url{https://platform.openai.com/docs/models/gpt-4o}} to construct question-answer pairs, and given its excellent visual comprehension, we take the remote sensing image and the corresponding scene category (provided by Million-AID and fMoW) as part of the prompt to generate questions and answers that are closely related to the image. An example of such a prompt is illustrated in Appendix~\ref{sec:appendix_ann}. In addition, following~\cite{lai2024lisa}, to make the questions and answers diverse, we adapt GPT-3.5 to rephrase the instructional questions and answers, as shown in Appendix Figure~\ref{fig:data_gen2}.

\noindent
\textbf{Object Mask Labeling.} Different from previous referring and reasoning segmentation datasets (which use off-the-shelf masks or bounding boxes), we annotate images from scratch. Specifically, we employ multiple experts in remote sensing and vision, assign each expert a few hundred images to annotate, and cross-validate the annotations after they are completed. For simple targets (\eg, lake), SAM-H~\cite{kirillov2023segment} is used to assist in annotation; for complex targets (\eg, wind turbine), each point of the polygon is finely marked. A description of mask quality is provided in Appendix~\ref{sec:appendix_ann}.


\noindent
\textbf{Dataset Statistics.}
The EarthReason dataset is partitioned into training, validation, and testing sets, comprising 2,371, 1,135, and 1,928 images, respectively. In the training set, each image is annotated with an average of six questions and three corresponding answers. The average question length is 20.86 words, while the average answer length is 26.76 words. To assess the model’s generalization capability, several semantic categories are deliberately reserved for the validation and test sets, ensuring they remain unseen during training. Additional dataset details are provided in the Appendix~\ref{sec:appendix_ann_stat}.

\section{Baseline Geospatial Pixel Reasoning Method---SegEarth-R1}

\begin{figure}[t]
  \centering
   \includegraphics[width=1.0\linewidth]{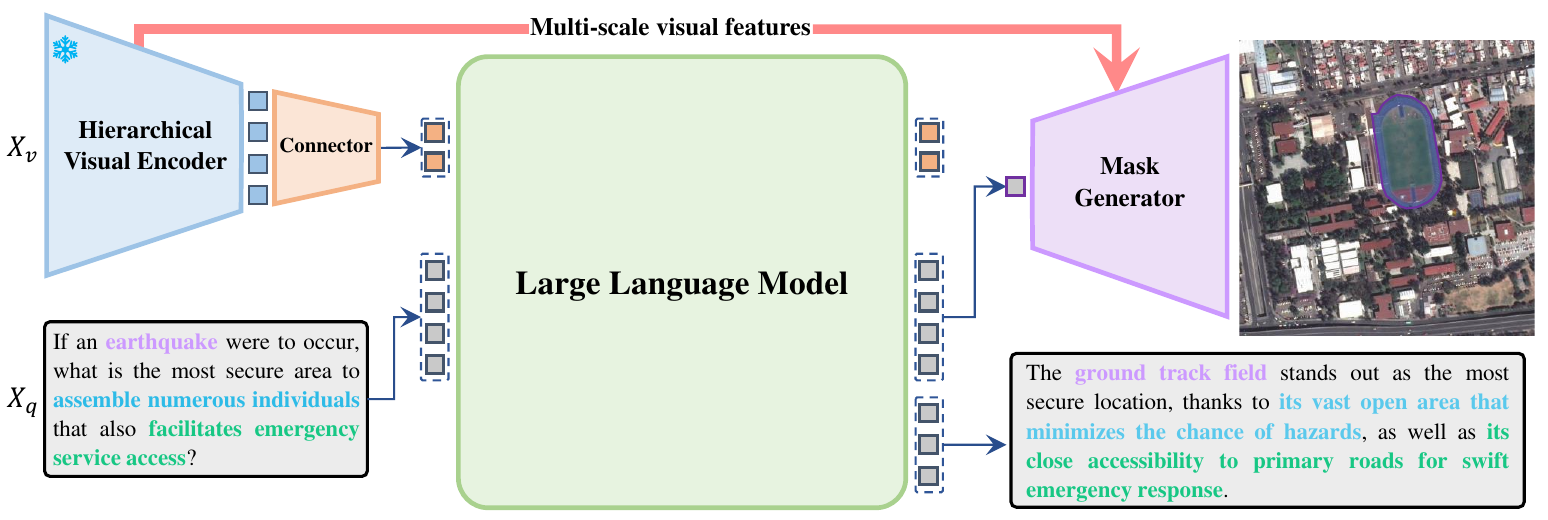}
   \caption{\small Overview of the proposed SegEarth-R1 architecture. Given an image $X_v$ and a text description $X_q$, a hierarchical visual encoder and a proposed connector are used to extract and compress visual tokens. Then, the visual tokens \includegraphics[scale=0.004,valign=c]{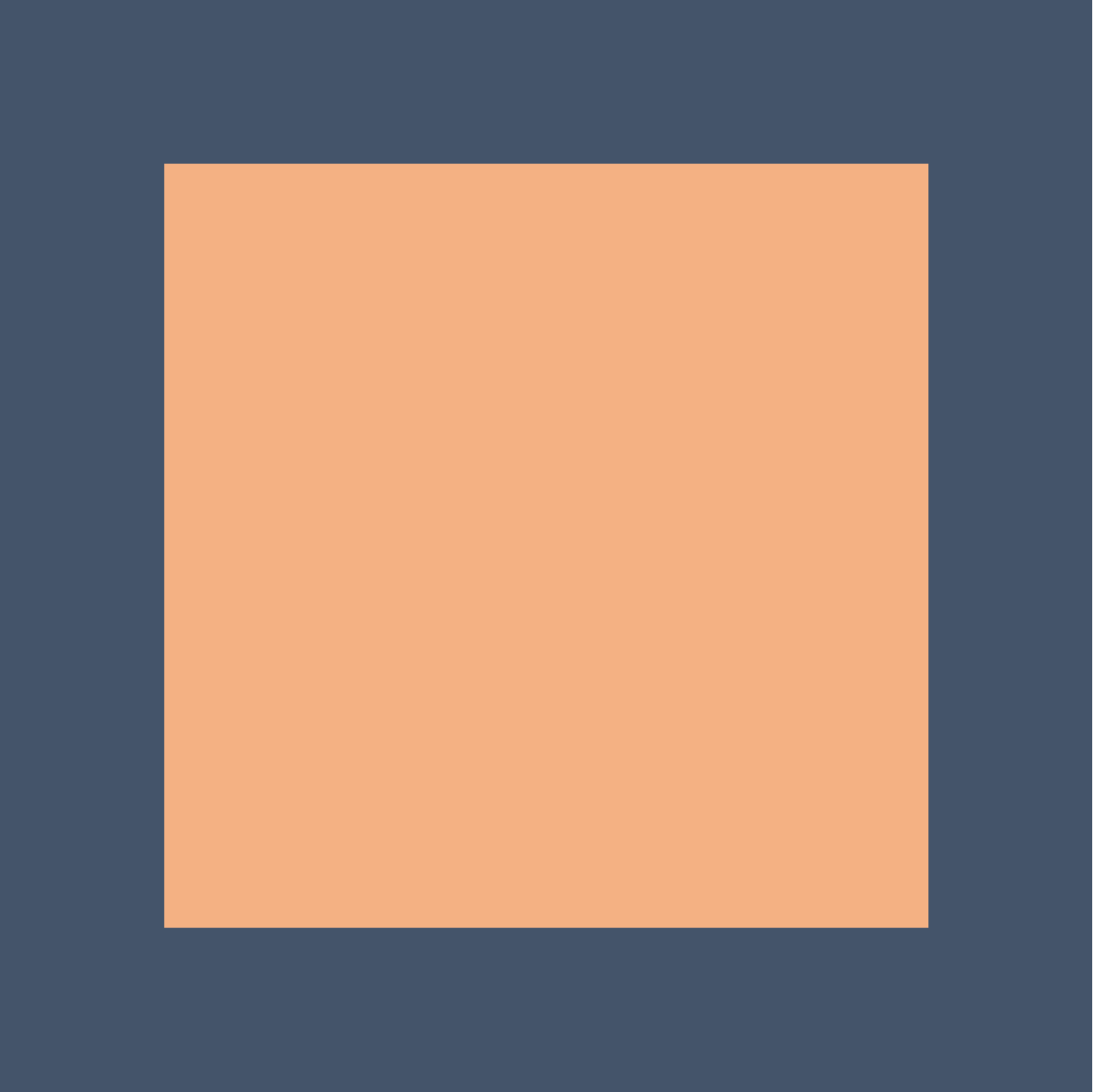} and description embeddings \includegraphics[scale=0.004,valign=c]{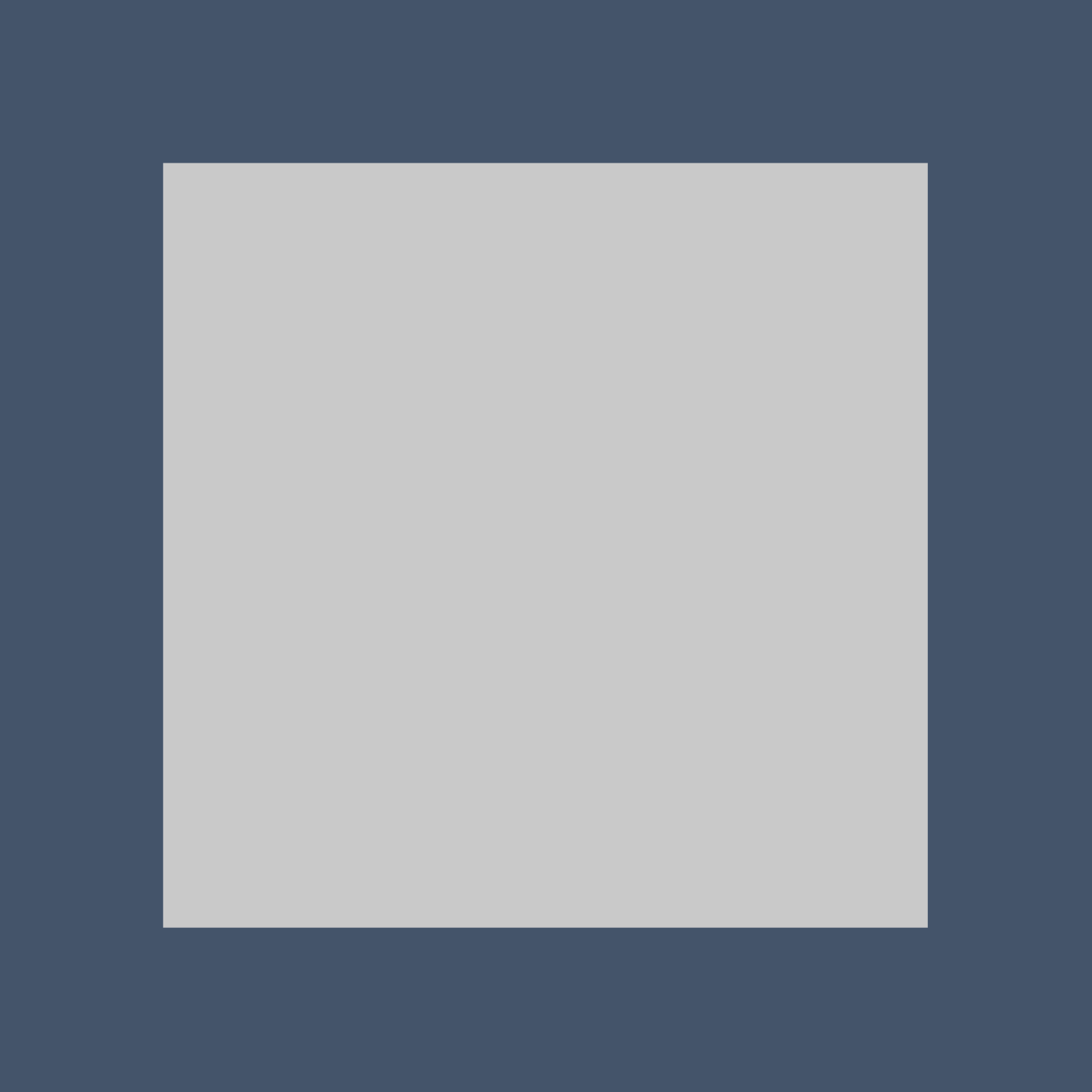} are fed into an LLM for instruction interpretation and semantic correlation. Finally, description embeddings are directly mapped to the query vector and used for spatial correlation and segmentation mask generation.}
   \label{fig:method}
   \vspace{-1em}
\end{figure}

Compared with natural images, remote sensing images exhibit distinctive characteristics that demand specialized architectural designs for pixel-wise geospatial reasoning. In this work, we propose SegEarth-R1, a simple yet powerful baseline for geospatial pixel reasoning that effectively harnesses LLM capabilities while incorporating domain-specific adaptations. As illustrated in Figure~\ref{fig:method}, our architecture comprises three core parts: A visual encoder for image feature extraction, an LLM for instruction interpretation and semantic correlation, and a mask generator for spatial correlation and mask prediction. Each part incorporates critical design considerations to address the unique challenges of remote sensing images.

\subsection{Hierarchical Visual Encoder}

Satellite and aerial targets present two critical challenges: (1) extreme scale variations ranging from sub-meter objects to kilometer-scale geographical formations~\cite{rolf2024mission}, and (2) densely distributed small objects requiring high-resolution analysis~\cite{li2024segearth}. Conventional ViT-based encoders adopted in MLLMs~\cite{lai2024lisa, yang2023lisa++, kirillov2023segment, xia2024gsva} (\eg, image encoder in CLIP~\cite{radford2021learning} and SAM~\cite{kirillov2023segment, ravi2024sam}) prove suboptimal due to their fixed-scale feature extraction and information compression through aggressive patch merging. To alleviate these limitations, following~\cite{zhang2024psalm}, SegEarth-R1 employs a Swin Transformer~\cite{liu2021swin} backbone enhanced with progressive feature hierarchy construction. This architecture generates multi-scale feature maps $v_h, h\in [1,4]$ at {1/4, 1/8, 1/16, 1/32} of the original resolution through controlled downsampling operations, preserving high-resolution details for small objects while capturing contextual semantics at deeper layers.


\subsection{Large Language Model and Input Schema}

SegEarth-R1 adopts the MLLM paradigm~\cite{liu2023visual, li2023blip} by jointly embedding visual tokens and textual instructions into a unified LLM input space for multimodal reasoning. Unlike natural images, remote sensing data exhibits ultra-high-resolution coverage~\cite{ji2023ultra, wang2025xlrs}, posing computational challenges when processed through billion-level LLMs. Therefore, we expect to compress the visual token to alleviate the computational cost and make only simple semantic correlations in LLM.

\subsubsection{Visual Token}

\textbf{Redundancy Analysis.}
Image redundancy quantifies the proportion of compressible, non-informative data within an image. To investigate the feasibility of aggressive visual token compression for remote sensing images, we conduct a redundancy analysis from dual perspectives: pixel-level statistical redundancy and spatial structural redundancy.


\begin{itemize}[leftmargin=8pt]
\item According to information theory~\cite{shannon1948mathematical}, entropy measures the average uncertainty or information content of an image, while the maximum entropy corresponds to the idealized scenario where pixel values are uniformly distributed (\ie, no redundancy). Thus, from the entropy perspective, the image redundancy can be defined as~\cite{gonzales1987digital}:
\begin{equation}
\begin{aligned}
  R_e = 1 - \frac{-\sum_{l=0}^{L-1} p(l) \log _{2} p(l)}{\log _{2} L},
  \label{eq:entropy}
\end{aligned}
\end{equation}
where $L$ denotes the number of distinct intensity levels (\eg, $L=256$ for an 8-bit grayscale image), and $p(l)$ denotes the probability mass function of the pixel intensity value $l$.
\end{itemize}

\begin{itemize}[leftmargin=8pt]
\item Beyond pixel-level statistical redundancy, structural self-similarity reflects spatial redundancy caused by repetitive patterns (\eg, textures, geometric features). To quantify this, we leverage the Structural Similarity Index Matrix (SSIM)~\cite{wang2004image} to measure inter-patch similarity. For an image partitioned into $N$ patches, the SSIM matrix $\mathbf{M} \in \mathbb{R}^{N \times N}$ is defined as:
\begin{equation}
\begin{aligned}
\mathbf{M}(i,j) = \frac{(2\mu_i\mu_j + C_1)(2\sigma_{ij} + C_2)}{(\mu_i^2 + \mu_j^2 + C_1)(\sigma_i^2 + \sigma_j^2 + C_2)}, \quad \forall i,j \in {1, ..., N}
\label{eq:ssim}
\end{aligned}
\end{equation}
where $\mu_{i}$, $\sigma_{i}$ denote the mean and variance of the $i$-th patch, $\sigma_{ij}$ is the covariance between patches $i$ and $j$, and $ C_{1}$, $C_{2}$ are stability constants. Then, the structural self-similarity redundancy $R_{s}$ is derived by averaging off-diagonal elements of $\mathbf{M}$:
\begin{equation}
\begin{aligned}
R_s = \frac{1}{N(N-1)} \sum_{i \neq j} \mathbf{M}(i,j).
\label{eq:ssim_redundancy}
\end{aligned}
\end{equation}
\end{itemize}

\begin{wrapfigure}{r}{0.6\textwidth} 
  \centering
  \vspace{-5pt}
  \subfloat[pixel-level redundancy]{\label{fig:redundancy_a}
    \includegraphics[width=0.28\textwidth]{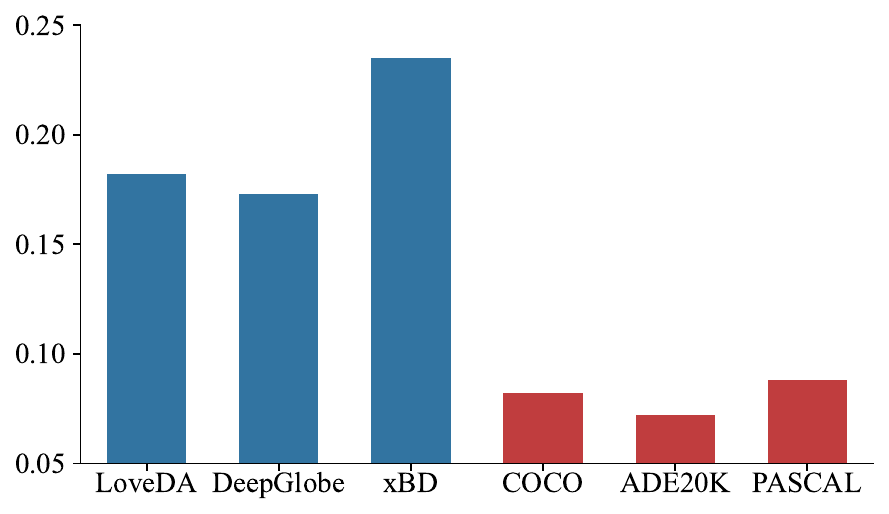}
  }
  \subfloat[spatial structure redundancy]{\label{fig:redundancy_b}
    \includegraphics[width=0.28\textwidth]{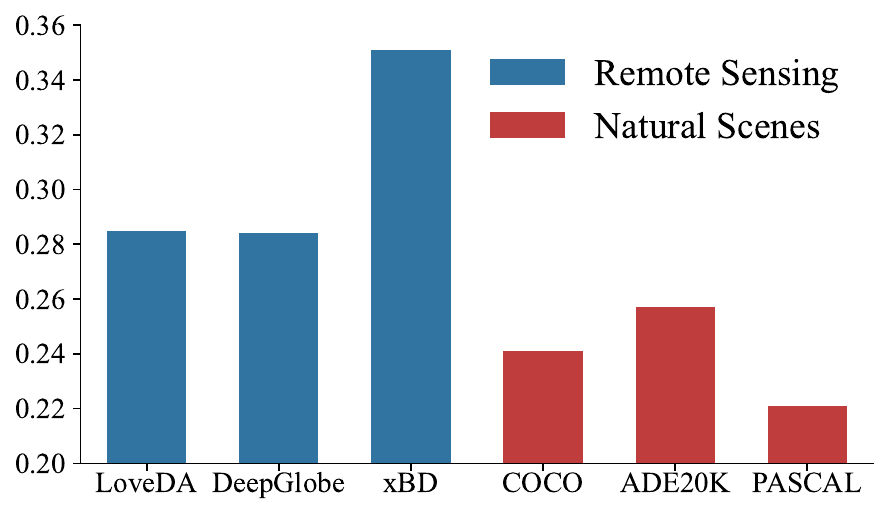}
  }
  \caption{\small Redundancy analysis of remote sensing datasets and natural images, and the former exhibits higher redundancy.}
  \label{fig:simg_pair}
  \vspace{-1em}
\end{wrapfigure}

We evaluate six benchmark datasets spanning natural images (COCO~\cite{caesar2018coco}, ADE20K~\cite{zhou2017scene}, PASCAL~\cite{everingham2010pascal}) and remote sensing images (LoveDA~\cite{wang2021loveda}, DeepGlobe~\cite{demir2018deepglobe}, xBD~\cite{gupta2019xbd}) for redundancy analysis. As shown in Figure~\ref{fig:simg_pair}, our analysis reveals two critical findings: 1) Remote sensing images demonstrate 1.9$\sim$3.3$\times$ higher entropic redundancy than natural images, indicating greater pixel-level compressibility. 2) The average self-similarity for remote sensing data exceeds natural images by 42.6\%, confirming the higher prevalence of repetitive textures and geometric patterns. This insight justifies aggressive token compression for semantic-level comprehension in remote sensing images.

\textbf{Token Compression Connector.}
In modern MLLM, connectors such as Q-Former~\cite{li2023blip} and MLP~\cite{liu2023visual} are designed to transform visual tokens into a multi-modal space. However, some works~\cite{cha2024honeybee, yao2024deco} point out that Q-Former may lead to loss of vision information and is difficult to train. Therefore, in SegEarth-R1, we follow the MLP connector fashion in LLaVA~\cite{liu2023visual} and use a simple but effective connector, \ie, stacked convolutional blocks and Layer Normalization (LN). Here, convolutional blocks are used for spatial down-sampling to compress the size of the feature map, and LN is used to stabilize cross-modal training. Specifically, our connector can be formulated as:
\begin{equation}
\begin{aligned}
v_{out}=(Conv \circ LN)^{d}(v_4),
\label{eq:connector}
\end{aligned}
\end{equation}
where $\circ$ denotes the function composition operator, and $d$ denotes the number of stacked layers.

\subsubsection{Text Instruction}

Although the instructions involved in geospatial pixel reasoning are implicit and contain more words than referring segmentation, they still maintain the same data format. Therefore, it is easy to convert them into question-answer pairs using a template like ``\textbf{USER}: This is an image <IMAGE>, please doing geospatial pixel reasoning according to the following instruction: <DESCRIPTION>. \textbf{ASSISTANT}: <ANSWER>''. For referring segmentation task, the task name in the instruction is changed to ``referring segmentation''.

\subsection{Mask Generation with Spatial Correlation}

Some recent LLM-based segmentation models~\cite{zhang2024psalm, wei2024instructseg} use Mask2Former~\cite{cheng2022masked} paradigm as mask generator. They use $T$ learnable mask tokens (typically $T$=100) as queries in the transformer decoder to generate $T$ candidate masks with corresponding scores, which are then assigned to the description embeddings\footnote{embeddings of <DESCRIPTION> in text instruction.} by bipartite matching. Unlike the reasoning segmentation of natural images, which is more inclined to make inferences based on the attributes of the object itself. In geospatial pixel reasoning, the model must understand and extrapolate more based on the spatial layout and inter-object correlations within the image (\eg, identifying earthquake evacuation zones in Figure~\ref{fig:method}, which requires analyzing topological relationships between roads and buildings). In addition, we posit that the mask query mechanism~\cite{cheng2021per, cheng2022masked} is inflexible and redundant in language-guided segmentation, and we only need to generate variable numbers of masks based on the instruction. Motivated by the above, we propose to directly use description embedding as the query for the mask generator and explicitly associate it with the image spatial features.

\begin{wrapfigure}{r}{0.25\textwidth}
    \vspace{-15pt}
    \centering
    \includegraphics[width=\linewidth]{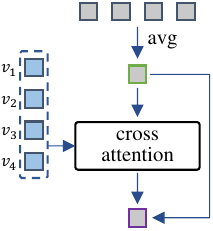} 
    \vspace{-10pt}
    \caption{\small $D$-Projector.}
    \label{fig:pool}
    \vspace{-10pt}
\end{wrapfigure}

The length of description embeddings varies according to the user's instruction, whereas in our geospatial pixel reasoning or referring segmentation setting, the segmentation result can be represented by a single binary mask. Therefore, we introduce a description projection module ($D$-Projector), which converts the whole description into a single vector as shown in Figure~\ref{fig:pool}. Specifically, the description embeddings are averaged into a global vector, which is then interacted with the flattened multi-scale visual features via a cross-attention operation, and a skip-connection and linear layer maps it into the query vector. Following this, the query vector is fed into the Transformer decoder of Mask2Former, which is composed of stacked masked attention, self-attention and FFN. Notably, since the mask query mechanism has been removed, the number of generated masks is the same as the number of queries, and therefore, score prediction and bipartite matching are also not required. Finally, the predicted mask is supervised with a linear combination of focal loss~\cite{lin2017focal} and dice loss~\cite{milletari2016v}.

\section{Experiments}

\subsection{Settings}

\noindent
\textbf{Datasets and Tasks.} In addition to the geospatial pixel reasoning on EarthReason, we assess the basic geolocation capability with the explicit short query of our model. We employed two benchmark referring image segmentation datasets: RefSegRS~\cite{yuan2024rrsis} and RRSIS-D~\cite{liu2024rotated}. These datasets contain 14 and 20 semantic categories, respectively, with textual descriptions primarily focusing on direct visual attributes such as orientation, color, and size. All models are trained on their own training set and evaluated on their validation and testing sets.


\noindent
\textbf{Evaluation metrics.} Following~\cite{lai2024lisa}, we adopt two evaluation metrics: \textbf{gIoU} (per-image IoU average) and cIoU (cumulative intersection over union). The latter is preferred because of its stability.

\noindent
\textbf{Network Architecture.} Unless otherwise specified, SegEarth-R1 use phi-1.5 (1.3B)~\cite{li2023textbooks} as the LLM, and adopt the Swin-B as the visual encoder. The token compression connector is configured with a layer number $d=2$. The mask generator follows the Mask2Former architecture, but removes mask tokens as mentioned above.

\noindent
\textbf{Implementation details.} During training, we use bf16 precision and freeze the visual encoder. The LLM is initialized from Phi-1.5, while both the Swin-B encoder and the mask generator are initialized with pretrained weights from Mask2Former. All images are resized to $1024 \times 1024$, maintaining the original aspect ratio by padding the shorter side. We adopt the AdamW optimizer with an initial learning rate of $1 \times 10^{-4}$, cosine learning rate schedule, and no weight decay. A uniform batch size of 16 is used across datasets, with training steps set to 7,610 (RRSIS-D), 5,400 (RefSegRS), and 2,220 (EarthReason). All experiments are conducted on two NVIDIA A100 80GB GPUs.

\subsection{Geospatial Pixel Reasoning Results}

\begin{wraptable}{l}{0.7\textwidth}
\vspace{-1em}
    \caption{\small Geospatial pixel reasoning results among SegEarth-R1 (ours) and previous related works.}
    \label{tab:earthreason}
    \centering
    \scalebox{0.75}{
    \begin{tabular}{@{}l|c|c|cc|cc@{}}
    \toprule[1pt]
    \multirow{2}{*}{Method} & \multirow{2}{*}{Visual Encoder} & \multirow{2}{*}{LLM Type} & \multicolumn{2}{c|}{cloU} & \multicolumn{2}{c}{gloU} \\
    & & & Val & Test & Val & Test \\
    \midrule[1pt]
    LISA~\cite{lai2024lisa} & CLIP-L & Vicuna-7B~\cite{chiang2023vicuna} & 57.39 & 59.10 & 61.04 & 60.88 \\
    PixelLM~\cite{ren2024pixellm} & CLIP-L & Vicuna-7B~\cite{chiang2023vicuna} & 57.79 & 59.22 & 57.94 & 60.01 \\
    PSALM~\cite{zhang2024psalm} & Swin-B & phi-1.5 (1.3B)~\cite{li2023textbooks} & 62.03 & 64.61 & 66.61 & 68.30 \\
    \textit{SegEarth-R1} & Swin-B & phi-1.5 (1.3B)~\cite{li2023textbooks} & \textbf{64.13} & \textbf{68.25} & \textbf{68.60} & \textbf{70.75} \\
    \bottomrule[1pt]
\end{tabular}}
\vspace{-1em}
\end{wraptable}

We conduct a comparative evaluation of SOTA LLM-based methods and SegEarth-R1 on the EarthReason dataset. As shown in Table~\ref{tab:earthreason}, all models are trained solely on the training split of EarthReason to ensure a fair comparison. LISA and PixelLM demonstrate comparable performance; however, despite leveraging larger LLM or MLLM, the quality of their predicted segmentation masks remains suboptimal. This can be primarily attributed to their reliance on CLIP as the visual encoder, which tends to diminish the representation of small-scale geospatial targets. As one of the baselines of SegEarth-R1, PSALM achieves notable improvements over LISA and PixelLM. Nevertheless, PSALM does not adequately incorporate LLM-based segmentation and the Mask2Former paradigm, and lacks considerations for overhead images. SegEarth-R1 achieves the best results on both metrics surpassing PSALM by 3.64\% and 2.45\% on the test set. Importantly, SegEarth-R1 uses fewer visual tokens in LLM and reduces the number of queries in the mask generator, thus providing a lower inference cost.

\subsection{Referring Segmentation Results}

\begin{wraptable}{r}{0.55\textwidth}
    \vspace{-1em}
    \caption{\small Referring segmentation results among SegEarth-R1 and previous related works on RRSIS-D dataset.}
    \label{tab:RRSIS_D}
    \centering
    \scalebox{0.65}{
    \begin{tabular}{@{}l|c|cc|cc|cc@{}}
    \toprule[1pt]
    \multirow{2}{*}{Method} & & \multicolumn{2}{c|}{P@0.5} & \multicolumn{2}{c|}{cloU} & \multicolumn{2}{c}{gloU} \\
    & & Val & Test & Val & Test & Val & Test \\
    \midrule[1pt]
    \multicolumn{8}{@{}l}{\textbf{\textit{Traditional method:}}} \\
    RRN~\cite{li2018referring} & {\tiny CVPR'18} & 51.09 & 51.07 & 66.53 & 66.43 & 46.06 & 45.64 \\
    CSMC~\cite{ye2019cross} & {\tiny CVPR'19} & 55.68 & 55.32 & 69.39 & 69.39 & 48.85 & 48.54 \\
    LSCM~\cite{hui2020linguistic} & {\tiny ECCV'20} & 57.12 & 56.02 & 69.05 & 69.28 & 50.36 & 49.92 \\
    CMPC~\cite{huang2020referring} & {\tiny CVPR'20} & 57.93 & 55.83 & 69.22 & 69.39 & 50.41 & 49.24 \\
    BRINet~\cite{hu2020bi} & {\tiny CVPR'20} & 58.79 & 56.90 & 70.73 & 69.88 & 51.14 & 49.65 \\
    CMPC+~\cite{liu2021cross} & {\tiny TPAMI'20} & 59.19 & 57.65 & 70.14 & 68.64 & 51.41 & 50.24 \\
    LGCE~\cite{yuan2024rrsis} & {\tiny TGRS'24} & 68.10 & 67.65 & 76.68 & 76.34 & 60.16 & 59.37 \\
    RIS-DMMI~\cite{hu2023beyond} & {\tiny CVPR'23} & 70.40 & 68.74 & 77.01 & 76.20 & 60.72 & 60.12 \\
    LAVT~\cite{yang2022lavt} & {\tiny CVPR'22} & 69.54 & 69.52 & 77.59 & 77.19 & 61.46 & 61.04 \\
    RMSIN~\cite{liu2024rotated} & {\tiny CVPR'24} & 74.66 & 74.26 & 78.27 & 77.79 & 65.10 & 64.20 \\
    \midrule[1pt]
    \multicolumn{8}{@{}l}{\textbf{\textit{LLM-based method:}}} \\
    LISA~\cite{lai2024lisa} & {\tiny CVPR'24} & 27.07 & 24.51 & - & - & 27.84 & 26.78 \\
    PixelLM~\cite{ren2024pixellm} & {\tiny CVPR'24} & 33.46 & 28.81 & - & - & 33.89 & 31.65 \\
    NEXT-Chat~\cite{zhang2023next} & {\tiny arXiv'23} & 28.97 & 26.37 & - & - & 26.98 & 24.98 \\
    GeoGround~\cite{zhou2024geoground} & {\tiny arXiv'25} & 68.69 & 67.50 & - & - & 61.10 & 60.50 \\
    \midrule[1pt]
    \multicolumn{2}{@{}l|}{\model} & \textbf{78.62} & \textbf{76.96} & \textbf{78.92} & \textbf{78.01} & \textbf{67.56} & \textbf{66.40} \\
    \bottomrule[1pt]
    \end{tabular}}
\end{wraptable}

SegEarth-R1 also supports basic explicit language-guided segmentation. As shown in Table~\ref{tab:RRSIS_D}, we compare its performance with existing SOTA traditional methods (not based on LLM) as well as recent LLM-based methods. Notably, prior to SegEarth-R1, LLM-based methods consistently underperformed in comparison to traditional methods on the referring segmentation task. For instance, the advanced GeoGround~\cite{zhou2024geoground} lags behind RMSIN~\cite{liu2024rotated} by 3.7\% in terms of gIoU on the RRSIS-D dataset. In contrast, SegEarth-R1, as a universal LLM-based language-guided segmentation method, surpasses traditional methods on the referring segmentation task for the first time with a 2.2\% improvement. This result highlights the enhanced generalization capability and practical potential of SegEarth-R1. On the RefSegRS dataset, the improvement of SegEarth-R1 is more significant than the previous method, with an 8.33\% and 9.87\% improvement over RMSIN on the validation and testing sets, respectively, as listed in Table~\ref{tab:RefSegRS}.

\begin{table}[h]
    \caption{\small Referring segmentation results among SegEarth-R1 and previous related works on RefSegRS dataset.}
    \label{tab:RefSegRS}
    \centering
    \scalebox{0.66}{
    \begin{tabular}{@{}l|c|cc|cc|cc|cc|cc|cc|cc@{}}
    \toprule[1pt]
    \multirow{2}{*}{Method} & & \multicolumn{2}{c|}{P@0.5} & \multicolumn{2}{c|}{P@0.6} & \multicolumn{2}{c|}{P@0.7} & \multicolumn{2}{c|}{P@0.8} & \multicolumn{2}{c|}{P@0.9} & \multicolumn{2}{c|}{cloU} & \multicolumn{2}{c}{gloU} \\
    & & Val & Test & Val & Test & Val & Test & Val & Test & Val & Test & Val & Test & Val & Test \\
    \midrule[1pt]
    BRINet~\cite{hu2020bi} & {\tiny CVPR'20} & 36.86 & 20.72 & 35.53 & 14.26 & 19.93 & 9.87 & 10.66 & 2.98 & 2.84 & 1.14 & 61.59 & 58.22 & 38.73 & 31.51 \\
    LSCM~\cite{hui2020linguistic} & {\tiny ECCV'20} & 56.82 & 31.54 & 41.24 & 20.41 & 21.85 & 9.51 & 12.11 & 5.29 & 2.51 & 0.84 & 62.82 & 61.27 & 40.59 & 35.54 \\
    CMPC~\cite{huang2020referring} & {\tiny CVPR'20} & 46.09 & 32.36 & 26.45 & 14.14 & 12.76 & 6.55 & 7.42 & 1.76 & 1.39 & 0.22 & 63.55 & 55.39 & 42.08 & 40.63 \\
    CMSA~\cite{ye2019cross} & {\tiny CVPR'19} & 39.24 & 28.07 & 38.44 & 20.25 & 20.39 & 12.71 & 11.79 & 5.61 & 1.52 & 0.83 & 65.84 & 64.53 & 43.62 & 41.47 \\
    RRN~\cite{li2018referring} & {\tiny CVPR'18} & 55.43 & 30.26 & 42.98 & 23.01 & 23.11 & 14.87 & 13.72 & 7.17 & 2.64 & 0.98 & 69.24 & 65.06 & 50.81 & 41.88 \\
    EVF-SAM~\cite{zhang2024evf} & {\tiny Arxiv'24} & 57.77 & 35.17 & 37.59 & 22.34 & 16.24 & 9.36 & 4.87 & 2.86 & 1.86 & 0.39 & 59.61 & 55.51 & 46.98 & 36.64 \\
    CMPC+~\cite{liu2021cross} & {\tiny TPAMI'21} & 56.84 & 49.19 & 37.59 & 28.31 & 20.42 & 15.31 & 10.67 & 8.12 & 2.78 & 0.55 & 70.62 & 66.53 & 47.13 & 43.65 \\
    CARIS~\cite{liu2023caris} & {\tiny ACMMM'23} & 68.45 & 45.40 & 47.10 & 27.19 & 25.52 & 15.08 & 14.62 & 8.87 & 3.71 & 1.98 & 75.79 & 69.74 & 54.30 & 42.66 \\
    CRIS~\cite{wang2022cris} & {\tiny CVPR'22} & 53.13 & 35.77 & 36.19 & 24.11 & 24.36 & 14.36 & 11.83 & 6.38 & 2.55 & 1.21 & 72.14 & 65.87 & 53.74 & 43.26 \\
    LAVT~\cite{yang2022lavt} & {\tiny CVPR'22} & 80.97 & 51.84 & 58.70 & 30.27 & 31.09 & 17.34 & 15.55 & 9.52 & 4.64 & 2.09 & 78.50 & 71.86 & 61.53 & 47.40 \\
    RIS-DMMI~\cite{hu2023beyond} & {\tiny CVPR'23} & 86.17 & 63.89 & 74.71 & 44.30 & 38.05 & 19.81 & 18.10 & 6.49 & 3.25 & 1.00 & 74.02 & 68.58 & 65.72 & 52.15 \\
    LGCE~\cite{yuan2024rrsis} & {\tiny TGRS'24} & 90.72 & 73.75 & 86.31 & 61.14 & 71.93 & 39.46 & 32.95 & 16.02 & 10.21 & 5.45 & 83.56 & 76.81 & 72.51 & 59.96 \\
    RMSIN~\cite{liu2024rotated} & {\tiny CVPR'24} & 93.97 & 79.20 & 89.33 & 65.99 & 74.25 & 42.98 & 29.70 & 16.51 & 7.89 & 3.25 & 82.41 & 75.72 & 73.84 & 62.58 \\
    SegEarth-R1 & & \textbf{95.82} & \textbf{86.30} & \textbf{93.27} & \textbf{79.53} & \textbf{88.86} & \textbf{69.57} & \textbf{78.19} & \textbf{48.87} & \textbf{22.04} & \textbf{10.73} & \textbf{85.01} & \textbf{79.00} & \textbf{82.17} & \textbf{72.45} \\
    \bottomrule[1pt]
\end{tabular}}
\end{table}

\subsection{Ablation Study}

\begin{figure}[t]
  \centering
   \includegraphics[width=1.0\linewidth]{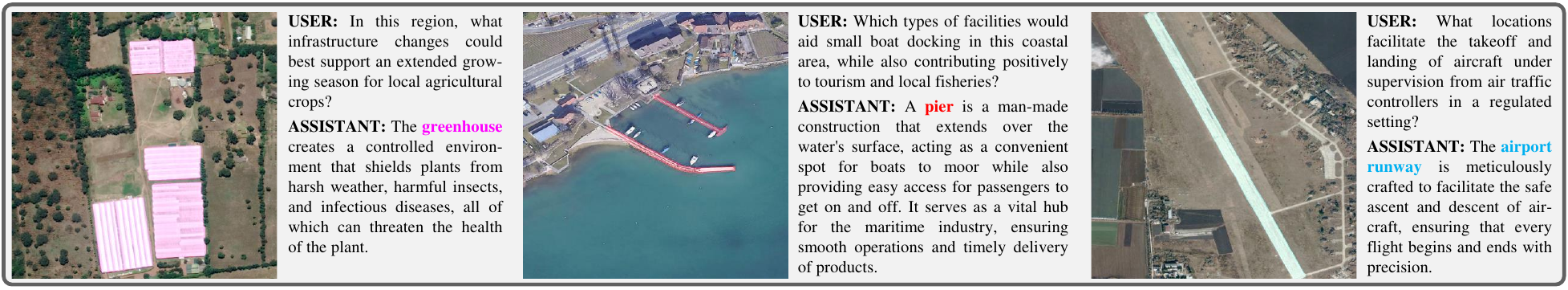}
   \caption{\small Qualitative Results of SegEarth-R1 on EarthReason. More results can be found in  Appendix~\ref{sec:appendix_examples}.}
   \label{fig:res}
   \vspace{-1em}
\end{figure}

\begin{minipage}{\textwidth}
\footnotesize
\begin{minipage}[t]{0.55\textwidth}
\makeatletter\def\@captype{table}
\caption{\small Ablation of SegEarth-R1 components on EarthReason: query description embedding (Query D.E.), description projector ($D$-Projector), token compression connector (T.C. Connector).}
\label{tab:ablation}
\centering
\scalebox{0.7}{
\begin{tabular}{@{}c|c|c|cc|cc@{}}
    \toprule[1pt]
    \multirow{2}{*}{Query D.E.} & \multirow{2}{*}{$D$-Projector} & \multirow{2}{*}{T.C. Connector} & \multicolumn{2}{c|}{cloU} & \multicolumn{2}{c}{gloU} \\
    & & & Val & Test & Val & Test \\
    \midrule[1pt]
    \ding{55} & \ding{55} & \ding{55} & 62.03 & 64.61 & 66.61 & 68.30 \\
    \Checkmark & \ding{55} & \ding{55} & 63.34 & 66.19 & 67.42 & 69.15 \\
    \ding{55} & \Checkmark & \ding{55} & 63.32 & 66.31 & 67.22 & 69.21 \\
    \ding{55} & \ding{55} & \Checkmark & 63.47 & 65.41 & 68.31 & 69.20 \\
    \Checkmark & \Checkmark & \ding{55} & 64.12 & 66.71 & \textbf{68.61} & 69.61 \\
    \Checkmark & \Checkmark & \Checkmark & \textbf{64.13} & \textbf{68.25} & 68.60 & \textbf{70.75} \\
    \bottomrule[1pt]
\end{tabular}}
\end{minipage}
\begin{minipage}[t]{0.45\textwidth}
\begin{minipage}[t][0.105\textheight][t]{\textwidth}
\makeatletter\def\@captype{table}
\caption{\small Ablation of LLM type on RRSIS-D.}
\label{table:ablation_llm}
\centering
\scalebox{0.75}{
\begin{tabular}{@{}l|cc|cc@{}}
    \toprule[1pt]
    \multirow{2}{*}{LLM Type} & \multicolumn{2}{c|}{cloU} & \multicolumn{2}{c}{gloU} \\
    & Val & Test & Val & Test \\
    \midrule[1pt]
    phi-1.5 (1.3B) & 78.92 & 78.01 & 67.56 & 66.40 \\
    phi-2 (2B) & \textbf{78.98} & \textbf{78.35} & \textbf{67.91} & \textbf{66.67} \\
    Qwen2.5 (0.5B) & 78.53 & 77.87 & 67.70 & 66.49 \\
    \bottomrule[1pt]
\end{tabular}}
\end{minipage}
\begin{minipage}[t][0.01\textheight][t]{\textwidth} 
\makeatletter\def\@captype{table}
\caption{\small Ablation of $d$ on EarthReason Val set.}
\label{table:ablation_d}
\centering
\scalebox{0.7}{
\begin{tabular}{@{}c|c|c||c|c|c@{}}
    \toprule[1pt]
    $d$ & \#Visual Token & gIoU & $d$ & \#Visual Token & gIoU \\
    \midrule[1pt]
    0 & 1024 & 68.28 & \textbf{2} & \textbf{64} & \textbf{68.60} \\
    1 & 256 & 68.47 & 3 & 16 & 68.22 \\
    \bottomrule[1pt]
\end{tabular}}
\end{minipage}
\end{minipage}
\end{minipage}

\noindent
\textbf{Components.}
We conduct ablation studies on the EarthReason dataset to evaluate the effectiveness of the novel components involved in SegEarth-R1. As listed in Table~\ref{tab:ablation}, the first row shows the results of the PSALM baseline. Each proposed component contributes to performance enhancement, yielding improvements ranging from 0.85\% to 0.9\%. The T.C. Connector and Query D.E. not only enhances performance but also reduces computational overhead. Further, the proposed components can be well coupled, and when they are all activated, \ie, complete SegEarth-R1, all metrics exhibit substantial gains over the baseline, confirming the effectiveness and compatibility of the proposed design. In fact, although these components are initially designed with remote sensing scenarios in mind, their underlying principles offer transferable insights applicable to general image understanding.

\noindent
\textbf{LLM Type.}
Given the limited scale of the dataset, we select some small LLM for comparation, as presented in Table~\ref{table:ablation_llm}. SegEarth-R1 demonstrates consistently high performance across different LLM, indicating the robustness and architectural stability of the overall framework. Notably, with Qwen2.5 (0.5B)~\cite{yang2024qwen2}, it still achieves competitive results, indicating its potential for edge deployment.

\noindent
\textbf{Layer Number of T.C. Connector.}
The layer number $d$ controls the number of visual tokens fed into the LLM. As shown in Table~\ref{table:ablation_d}, increasing token quantity does not improve performance. This observation aligns with our earlier analysis, suggesting that appropriate compression of visual tokens is beneficial for the global understanding of a remote sensing image. In SegEarth-R1, spatial correlations between the image and the instruction are primarily handled by the mask generator, while the LLM is only responsible for relatively semantic correlations. This division of labor allows for more efficient use of computational resources without compromising performance.

\section{Conclusion}

In this paper, we introduce geospatial pixel reasoning, a new task in remote sensing that requires models to infer segmentation masks from implicit natural language queries by reasoning over spatial context and domain knowledge. To enable research in this direction, we present EarthReason, the first large-scale benchmark dataset that emphasises complex reasoning scenarios. To address the distinct challenges inherent in remote sensing, we propose SegEarth-R1, a language-guided segmentation model that integrates a hierarchical visual encoder, an LLM for instruction parsing and semantic correlation, and a tailored mask generator designed for spatial correlation. Extensive experiments validate SegEarth-R1’s superiority, achieving SOTA performance on both geospatial pixel reasoning and referring segmentation tasks. This work pioneers the fusion of natural language reasoning with pixel-level geospatial analysis, offering transformative potential for applications like environmental monitoring and disaster response.

\bibliography{main}
\bibliographystyle{plain}

\newpage

\appendix

\section{Data}

\subsection{Annotation of EarthReason}
\label{sec:appendix_ann}

\begin{figure}[h]
  \centering
   \includegraphics[width=1.0\linewidth]{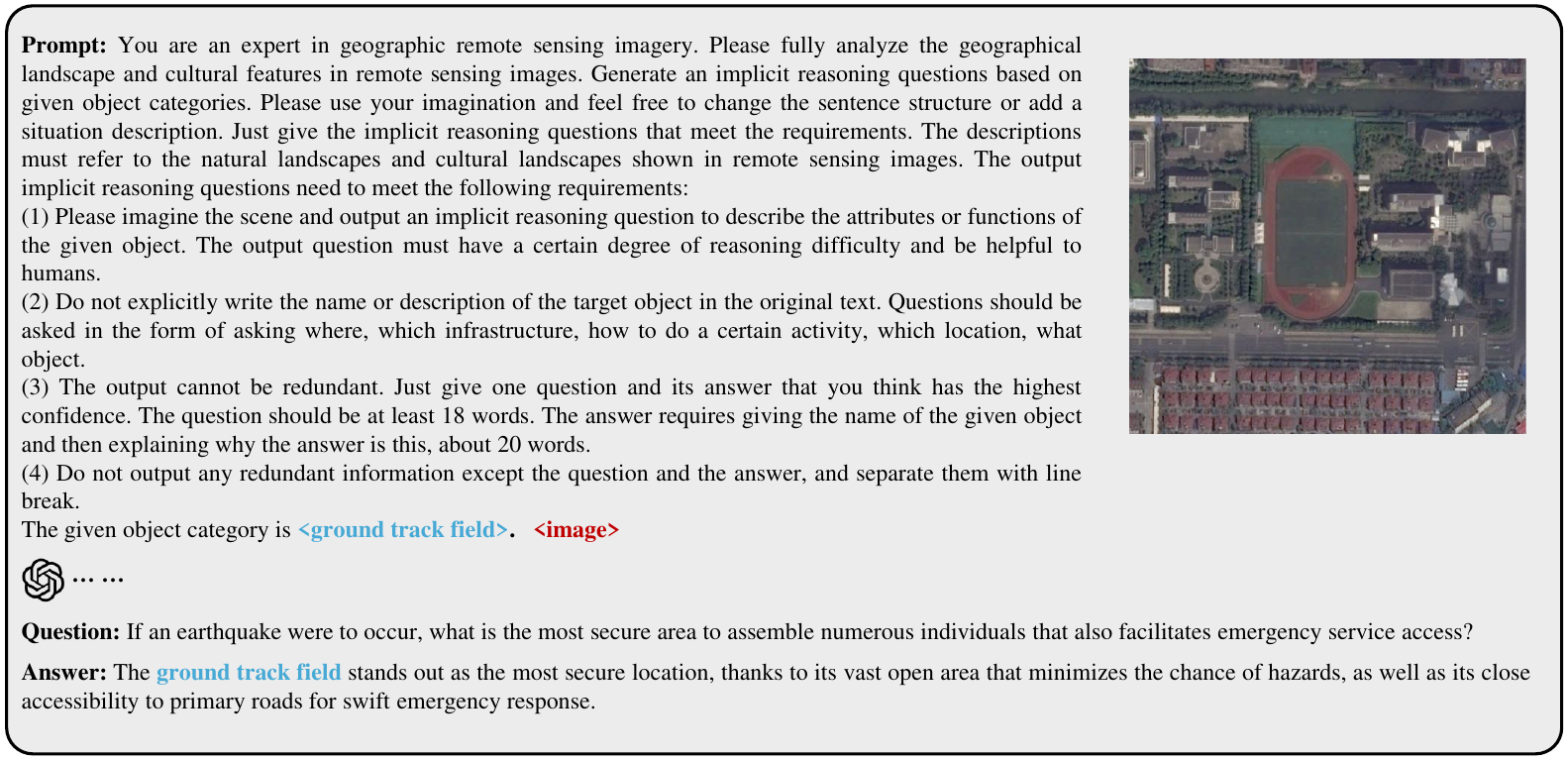}
   \caption{\small The illustration of the prompt construction process for generating question-answer pairs for geospatial pixel reasoning.}
   \label{fig:data_gen1}
\end{figure}

\begin{figure}[h]
  \centering
   \includegraphics[width=1.0\linewidth]{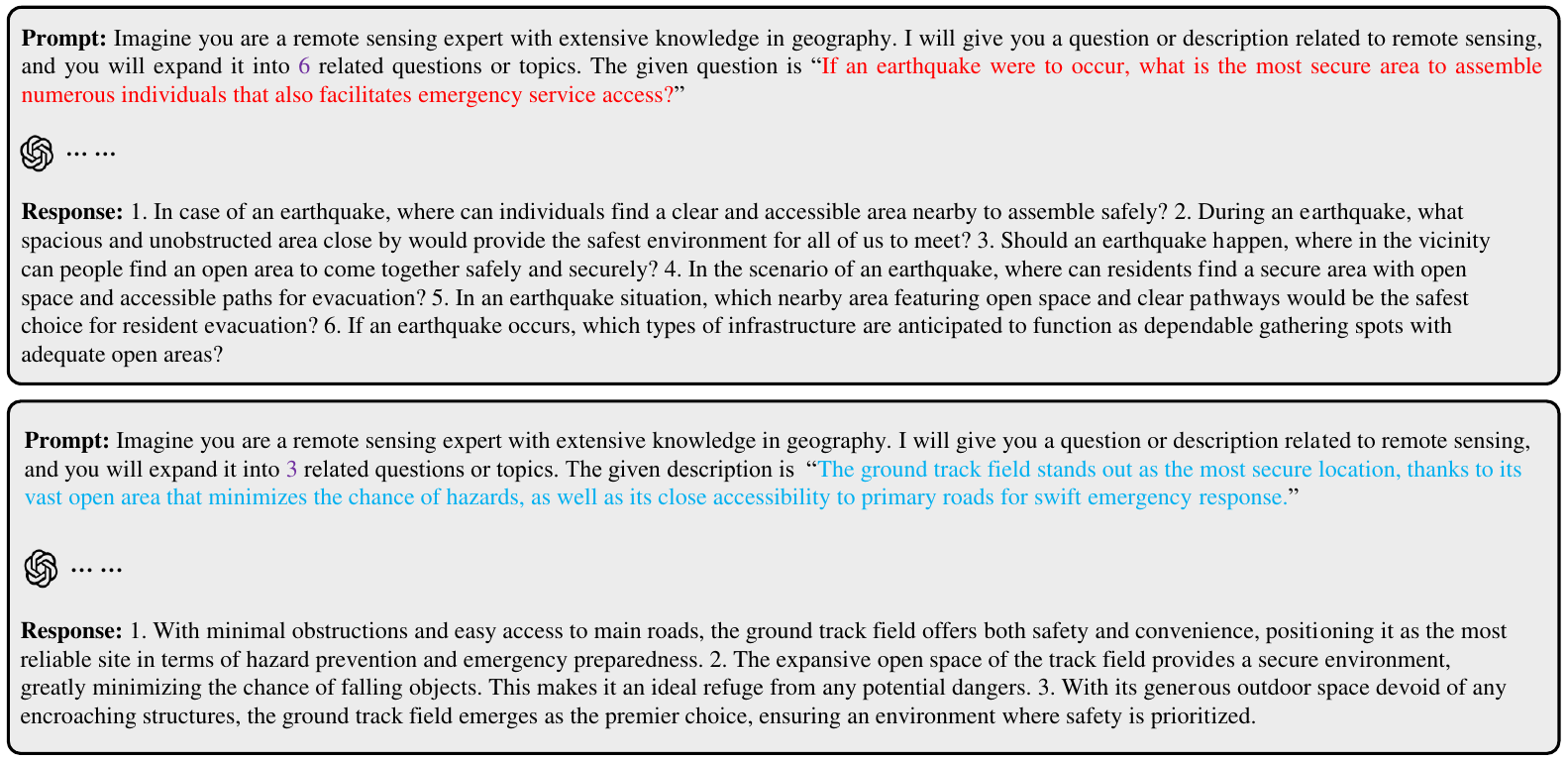}
   \caption{\small The illustration of the prompt construction process for expand question-answer pairs.}
   \label{fig:data_gen2}
\end{figure}

\begin{figure}[h]
  \centering
   \includegraphics[width=1.0\linewidth]{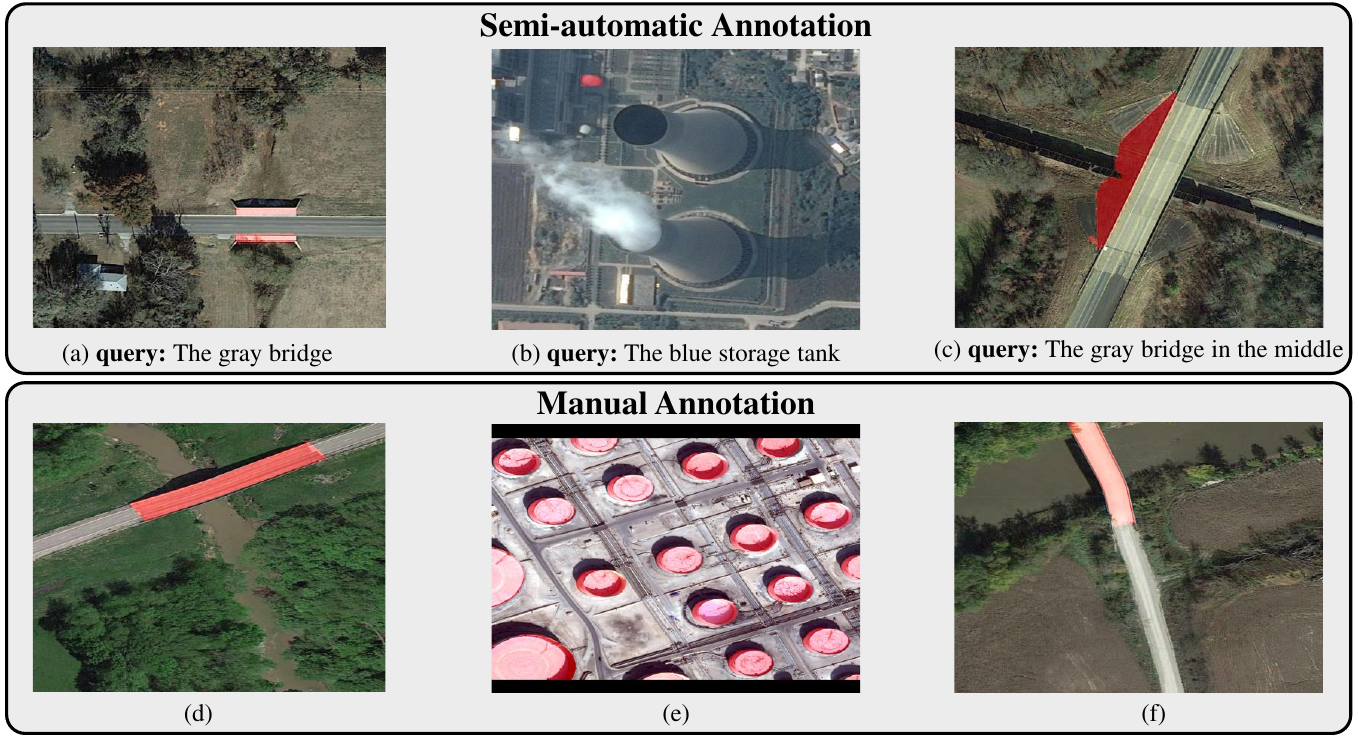}
   \caption{\small Comparison of annotation quality. (a), (b) and (c) are from RRSIS-D dataset, (d), (e) and (f) are from our EarthReason dataset.}
   \label{fig:dataset_compare}
\end{figure}

Each sample of the EarthReason benchmark consists of an image, a corresponding mask, and six reasoning queries along with their respective answers. Given that our metadata is derived from classification datasets, we employed GPT-4o and GPT-3.5 to generate textual annotations, and invited multiple remote sensing and vision experts to provide accurate and reliable mask annotations. Overall, our annotation process consists of the following three steps:

\begin{itemize}[leftmargin=1.5em]
    \item Step-1: To fully leverage the powerful multimodal capabilities and extensive geographic knowledge of GPT-4o, we carefully design the prompt, which is then provided alongside images and their corresponding category labels to generate a reasoning question–answer pair. The prompt is illustrated in Figure~\ref{fig:data_gen1}. 
    \item Step-2: To avoid homogeneous question–answer formats under a single prompt, we further employ the textual capabilities of GPT-3.5 to expand each generated question into six variations and each answer into three alternatives. The prompt used for this expansion is shown in Figure~\ref{fig:data_gen2}.
    \item Step-3: Unlike previous methods that rely on semi-automatic mask annotation based on off-the-shelf bounding boxes or masks, we invite multiple remote sensing vision experts to perform accurate and efficient mask annotation guided by the generated questions. To further improve annotation efficiency, we incorporate SAM-H as an auxiliary tool for some simple targets. Subsequently, we perform cross-validation of the annotation results and re-annotate the samples that do not meet the quality standards. As shown in Figure~\ref{fig:dataset_compare}, (a), (b), and (c), derived from the RRSIS-D dataset, illustrate the masks of semi-automatic annotation based on bounding boxes. (a) and (c) exhibit noticeable annotation errors, while in (b), the query does not align with the annotation. (d), (e), and (f) illustrate our high-quality manual annotations.
\end{itemize}


\subsection{EarthReason Statistics}
\label{sec:appendix_ann_stat}

\begin{figure}[h]
  \centering
   \includegraphics[width=1.0\linewidth]{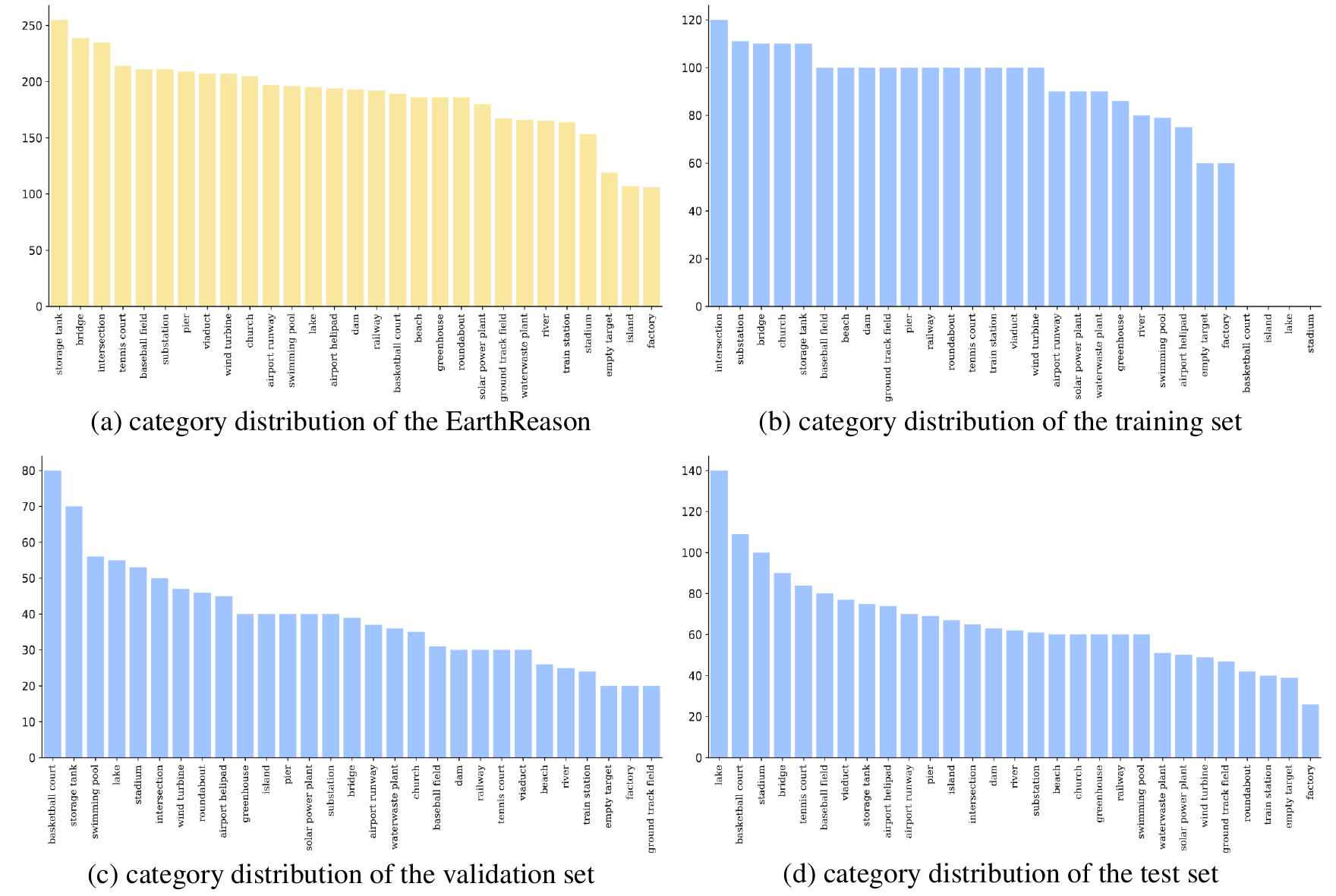}
   \caption{\small The category distribution of EarthReason.}
   \label{fig:category}
\end{figure}

The EarthReason benchmark comprises 28 categories, and the number of samples in each category is shown in Figure~\ref{fig:category} (a). It can be observed that the distribution of the 28 categories is relatively balanced. Figure~\ref{fig:category} (b), (c), and (d) illustrate the category distributions in the training, validation, and test sets, respectively. To evaluate the model's generalization capability, we specifically excluded four categories—``basketball court'', ``island'', ``lake'', and ``stadium''—from the training set. Moreover, we introduced 119 empty target samples to mitigate potential hallucinations of the model.

\section{Additional Implementation Details}
\label{sec:appendix_implementation}

\subsection{Details of Training Hyper-parameters}
Table~\ref{tab:hyperparam_training} presents the hyper-parameter settings used during the training of our model. For training on the referring segmentation datasets, we employ only focal loss and dice loss to supervise mask generation. In contrast, for training on geospatial pixel reasoning task, we additionally incorporate the cross-entropy loss from the large language model to supervise text answer generation.

\begin{table}[h]
\centering
\caption{The hyper-parameters for model training.}
\label{tab:hyperparam_training}
\begin{tabular}{ll}
\toprule
\textbf{Parameters} & \textbf{Value} \\
\midrule
Optimizer & AdamW \\
Learning Rate & $1 \times 10^{-4}$ \\
Batch Size & 16 \\
Number of Iteration & 7,610 / 5,400 / 2,220 \\
Learning Rate Schedule & Cosine Decay \\
Weight Decay & 0.0 \\
Warmup Ratio & 0.03 \\
$\beta_1$ & 0.9 \\
$\beta_2$ & 0.999 \\
Image Size & 1024 $\times$ 1024 \\
Image Processing & \begin{tabular}[t]{@{}l@{}}Resize long edge to 1024 \\ and padding short edge to 1024.\end{tabular} \\
\bottomrule
\end{tabular}
\end{table}

\section{Examples}
\label{sec:appendix_examples}

\subsection{More Qualitative Results on EarthReason}

\begin{figure}[h]
  \centering
   \includegraphics[width=1.0\linewidth]{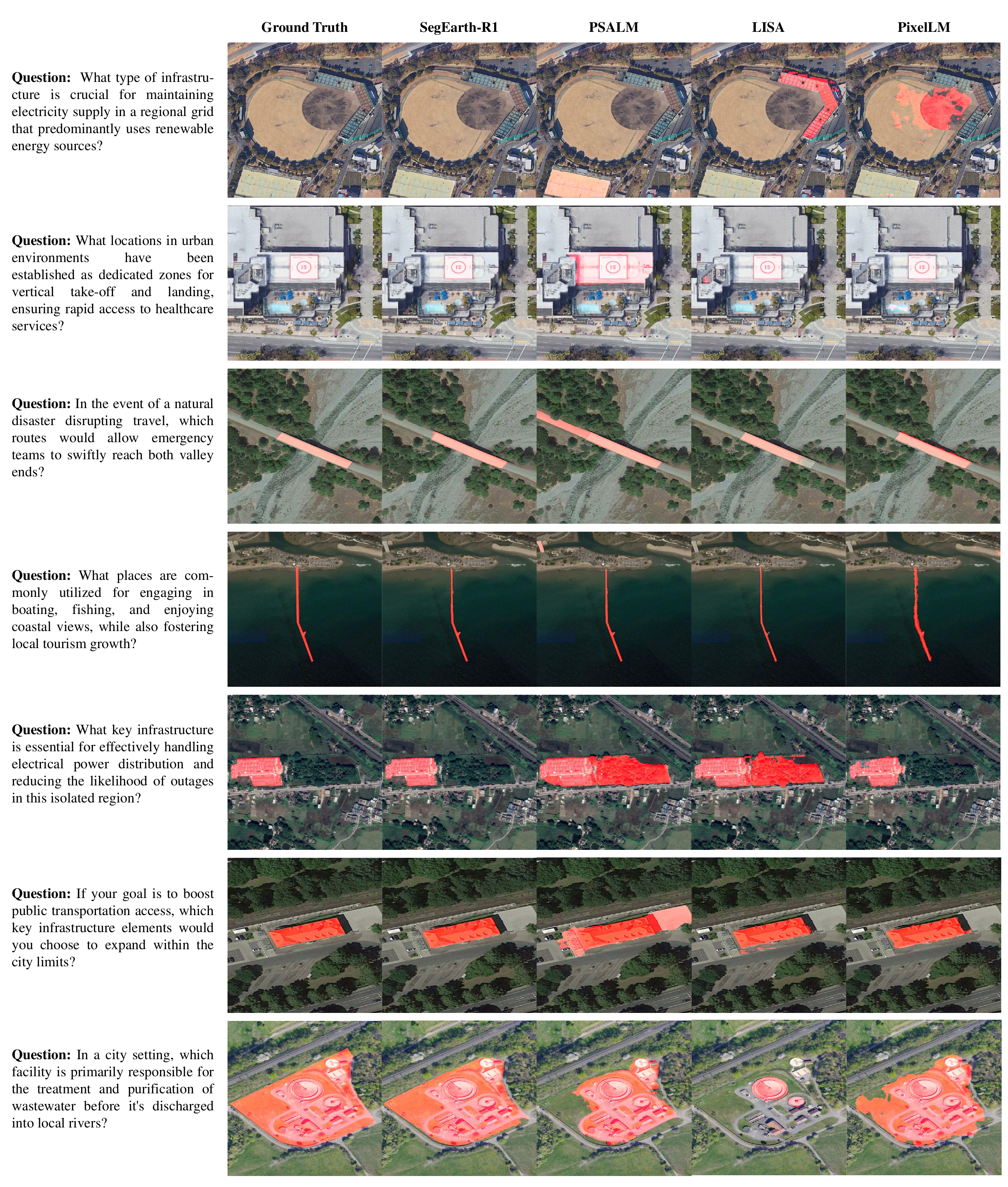}
   \caption{\small Comparison with other models on EarthReason.}
   \label{fig:vis_reason}
\end{figure}

Figure~\ref{fig:vis_reason} presents a comparison between SegEarth-R1 and other models on the EarthReason dataset. It can be observed that our model demonstrates a better understanding of long reasoning instructions and produces more accurate mask generation.

\subsection{More Qualitative Results on RRSIS-D}
\begin{figure}[h]
  \centering
   \includegraphics[width=1.0\linewidth]{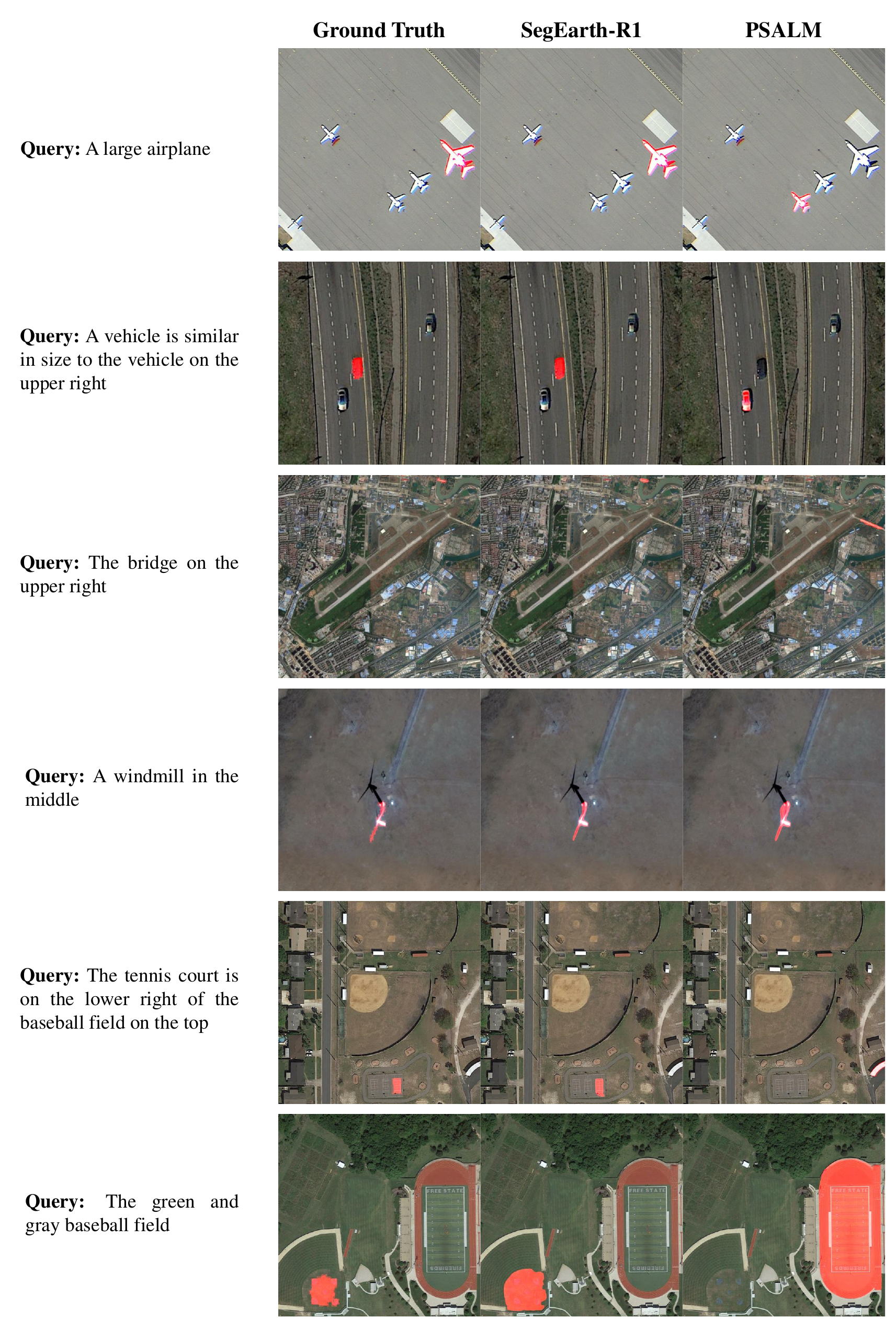}
   \caption{\small Comparison with PSALM on RRSIS-D.}
   \label{fig:vis_ref}
\end{figure}

Figure~\ref{fig:vis_ref} presents a comparison between SegEarth-R1 and PSALM on the RRSIS-D dataset. Our model demonstrates a better understanding of direct geographical attributes such as location, color, and size compared to PSALM. This improvement is attributed to the removal of indirect mask prediction using mask tokens, allowing semantic information (description embeddings) to directly interact with image features to generate masks.

\end{document}